\documentstyle[psfig,fleqn,aaai]{article}
\newcommand{\ignore}[1]{}
\newcommand{\comment}[1]{}

\begin{document}

\title{Evaluating Multilingual Gisting of Web Pages}
\author{
Philip Resnik \\ 
Department of Linguistics and \\
Institute for Advanced Computer Studies \\
University of Maryland \\
College Park, MD 20742 USA \\
{\em resnik@umiacs.umd.edu}
}
\date{}
\maketitle 

\begin{abstract}
We describe a prototype system for multilingual gisting of Web pages,
and present an evaluation methodology based on the notion of gisting
as decision support.  This evaluation paradigm is straightforward,
rigorous, permits fair comparison of alternative approaches, and
should easily generalize to evaluation in other situations where the
user is faced with decision-making on the basis of information in
restricted or alternative form.
\end{abstract}

\section{Introduction: Gisting as Decision Support} 
\label{sec:decision_support}

The word ``gisting'' has been used in a variety of settings.
Informally, it simply means ``getting the gist,'' that is, given some
information conveyed by natural language, understanding some
characteristic or important aspect of that information.

By definition, gisting is an activity in which the information taken
into account is less than the full information content available.  In
this paper, we take the view that there is another key aspect of
gisting that goes beyond simply selecting a subset of available
information, namely the goal of supporting {\em decision-making}.  In
an environment where human beings are attempting to gist radio
traffic, for example, radio operators need to decide whether or not to
route information to electronic warfare analysts \cite{dfactt1996}.
Accordingly, in order to evaluate a particular method for gisting, one
must examine the extent to which gisting supports a decision-making
task.

The focus of this paper is multilingual gisting on the World Wide Web,
with particular attention to developing a methodology for evaluating
multilingual gisting based on its role of decision support.  We see
such an evaluation methodology as important because, although the real
proof of any method is in how well it supports real users at their
real-world tasks, studying users in fully natural settings can be
difficult to organize, and, more important, two natural settings are
rarely similar enough to afford a fair comparison between alternative
approaches to the same task.  In order to address that problem, the
methodology we propose is applicable to a wide variety of tasks,
simple to carry out and, most important, defined in enough detail that
competing methods can be evaluated against the same set of data and
the results compared.

\section{Gisting for the Web: A Simple Prototype}
\label{sec:prototype}

\begin{figure}
\hbox{
\centerline{
\psfig{figure=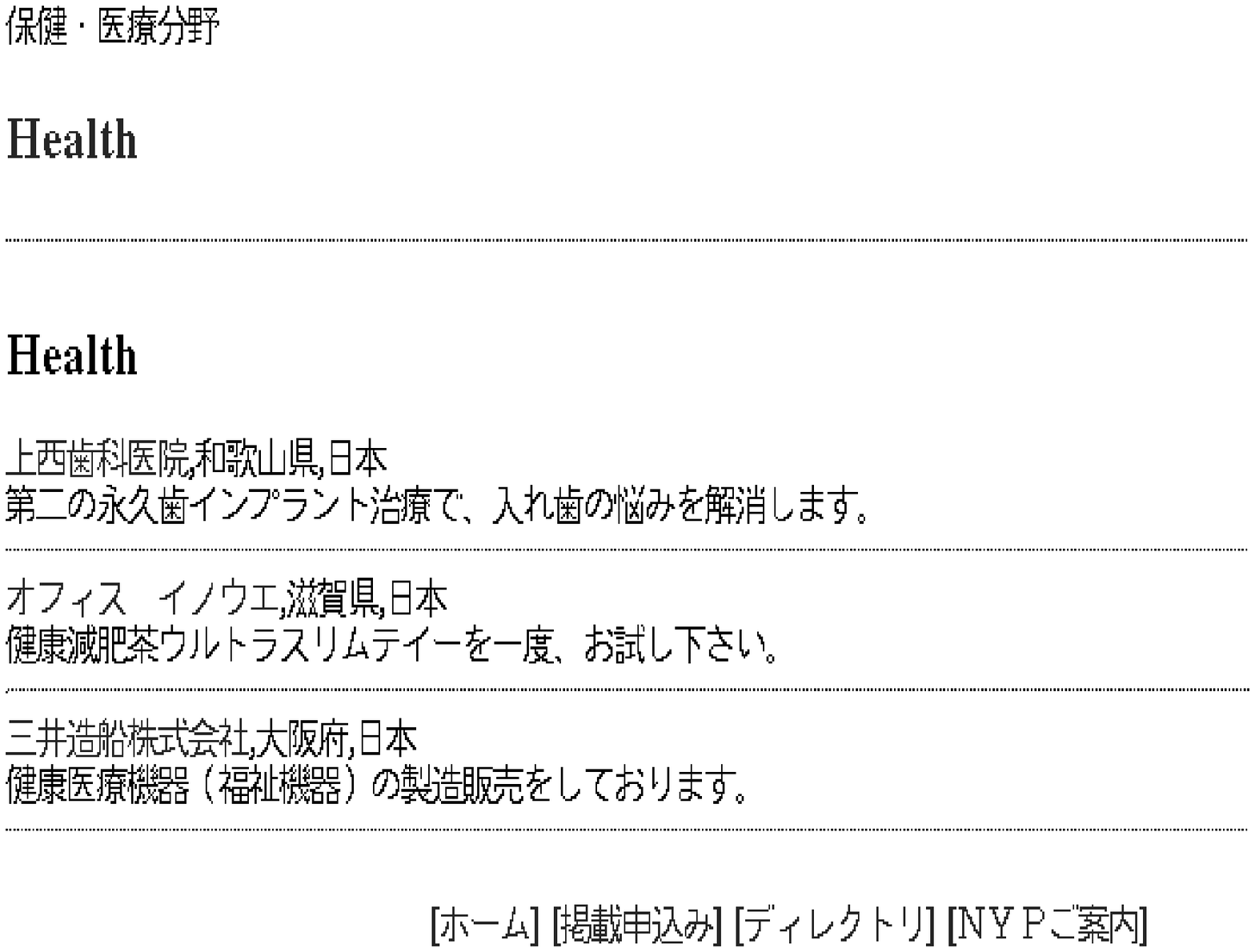,height=3.0in}}
}
\caption{Portion of a page from Nihongo Yellow Pages \label{fig:nyp_jp}}
\end{figure}

The motivation for this line of research can be described quite
simply.  Imagine that you are browsing the World Wide Web using your
favorite Web browser.  You click a link, or conduct a search, and find
yourself looking at the page illustrated in Figure~\ref{fig:nyp_jp}.
As it happens, you don't know a word of Japanese.  What are your
options? 
Is it worth finding a bilingual dictionary and looking up words on
this page? (And if so, which words?) Is it worth following links on
this page in hopes of finding something understandable?  (And if so,
which links?)  Is it worth bothering a nearby colleague who knows the
language and asking for a rough translation?  Is it worth going to the
time and expense of using an on-line service (e.g. The Global
Translation Alliance, http://www.aleph.com) to translate the page
completely?

In considering possible solutions to this scenario, we arrived at the
following principles.

\begin{quote}
{\bf Avoid full-scale machine translation.}  The user's problem would
certainly be solved by a fully automatic translation of the Web page
under consideration.  Unfortunately, the state of the art in high
quality machine translation is typically measured in words per minute
rather than pages per minute \cite{dorr_pc}, so even if it is {\em
possible\/} to obtain a translation for the page, the user is still
faced with the decision of whether or not it is worth sacrificing the
time to obtain it.

{\bf Keep the human in the loop.}  We see the problem scenario as an
opportunity for collaboration between person and machine, and in
particular an opportunity for the machine to facilitate the user in
doing things that people do well.  For example, people are capable of
disambiguating words almost effortlessly in context, although this is
a task at which computers currently perform quite poorly; therefore it
makes sense to have the computer present alternatives rather than
making disambiguation decisions for itself, unless such decisions can
be made with very high confidence.

{\bf Aim for extensibility.}  Our emphasis is on modular and
distributed design; for example, although we do not attempt to
disambiguate words in order to automatically select meaning
equivalents in the user's language, a disambiguation component could
easily be added to the system without wholesale changes in its design.
An ultimate target our efforts is the dissemination of application
programmer interfaces (APIs) that will make extensible infrastructure
available to the community at large.
\end{quote}

With those principles in mind, we implemented a prototype {\em gisting
proxy}, which assists users when confronted with a Web page in an
unknown language.\footnote{For the moment we are glossing over who
invokes the gisting proxy, and how.  In its full generality, this
proxy is part of a general design for a multilingual agent that is
aware of the user's linguistic knowledge and preferences, and goes
into action when it detects a situation where its capabilities might
assist the user.  For the current prototype, we have implemented a
gisting proxy HTTP server initiated by the user.}  When invoked for a
given Web page, the gisting proxy behaves as follows:
\begin{enumerate}
\item Convert the character encoding of the document into a standard
      encoding.
      \label{item:encoding}
\item Divide the Web page into structurally distinct pieces, using 
      HTML markup.  
      \label{item:markup}
\item For each piece:
  \begin{enumerate}
  \item Automatically identify the natural language in which this
        piece of text is written 
        \label{item:language_id}
  \item Invoke language-dependent word identification and
        normalization 
        \label{item:normalization}
  \item Look up each word in an on-line bilingual dictionary 
        \label{item:lookup}
  \item Present word-by-word glosses in the context of the original page
        \label{item:gloss}
  \end{enumerate}
\item Modify all links on the page so that further navigation from
      this point on will automatically go through the gisting proxy.
      \label{item:links}
\end{enumerate}

Step~\ref{item:encoding} is necessary because different character
encodings can be used for the same language, particularly in the case
of Asian languages (e.g. EUC-JP vs. Shift-JIS).  Normalization of
character encoding is necessary for consistency across components of
the system.

Step~\ref{item:markup} makes it possible to analyze documents
containing text in multiple languages.  Small sub-document units
(e.g. list items) motivate taking an approach to automated language
identification (Step~\ref{item:language_id}) that can work well even
when the strings to be identified are very short and cannot be relied
upon to contain function words \cite{dunning1994:language_id}.

Depending on the language, different measures must be taken in order
to identify words (Step~\ref{item:normalization}).  For example, in
many Asian languages words are typically not delimited by spaces, and
therefore automatic word segmentation is necessary \cite{juman}.  This
contrasts with Romance languages such as Spanish, where words are
generally delimited by spaces or punctuation but a small subset of the
lexical items in the language must be identified and separated out
(e.g. Spanish {\em damelo\/} = {\em da\/} + {\em me\/} + {\em
lo}).\comment{Gary comments that we should cite the Unicode 2.0 text
boundaries section and/or Taligent I18N web page (a.k.a. JDK 1.1 I18N
library support of word boundaries).}  In addition, some form of
normalization may need to be done as well.  For example, in order to
locate {\em da\/} in a Spanish-English translation lexicon it may be
necessary to look it up by its root form, {\em dar\/} (to give).

Word-by-word lookup and presentation in this system
(Steps~\ref{item:lookup} and~\ref{item:gloss}) resemble the direct
lexical approach to machine translation investigated and thoroughly
criticized in the 1960s \cite{alpac}.  Notably, however, the problem
attacked by those early efforts was one of full scale translation, not
gisting.  We would contend that with the rise of the World Wide Web,
those early solutions have finally found the right problem.

In the current prototype, presentation of the known-language glosses
for a word are guided by the results of the dictionary lookup.  At
present:
\begin{itemize}
\item If the unknown language word has a single gloss in the
      dictionary, show that gloss.
\item If the unknown language word has multiple glosses in the
      dictionary, show up to $n$ of them for some customizable
      parameter $n$ (currently $n=3$ by default), within parentheses
      and separated by commas.  For example, {\em (doctor's office, 
      clinic, dispensary)}.
\item If the unknown-language word is not found in the dictionary, then
  \begin{itemize}
  \item Show the unknown-language word itself, if the character set
        of the language is the same as a language the user knows
        (e.g. an unknown word in French would be shown to someone who
        knows English, since both use the Latin-1 character set).
  \item Show an ellipsis ($\ldots$) otherwise.
  \end{itemize}
\end{itemize}

This treatment of words not appearing in the dictionary follows the
general principle that users should be given information that might be
helpful --- such as possible cognates --- but minimally distracted by
unfamiliar scripts.  The present implementation reflects two extremes
for unknown words, namely presenting them as-is or leaving them out
entirely, but other strategies are possible.  

\begin{figure*}
\hbox{
\centerline{
\psfig{figure=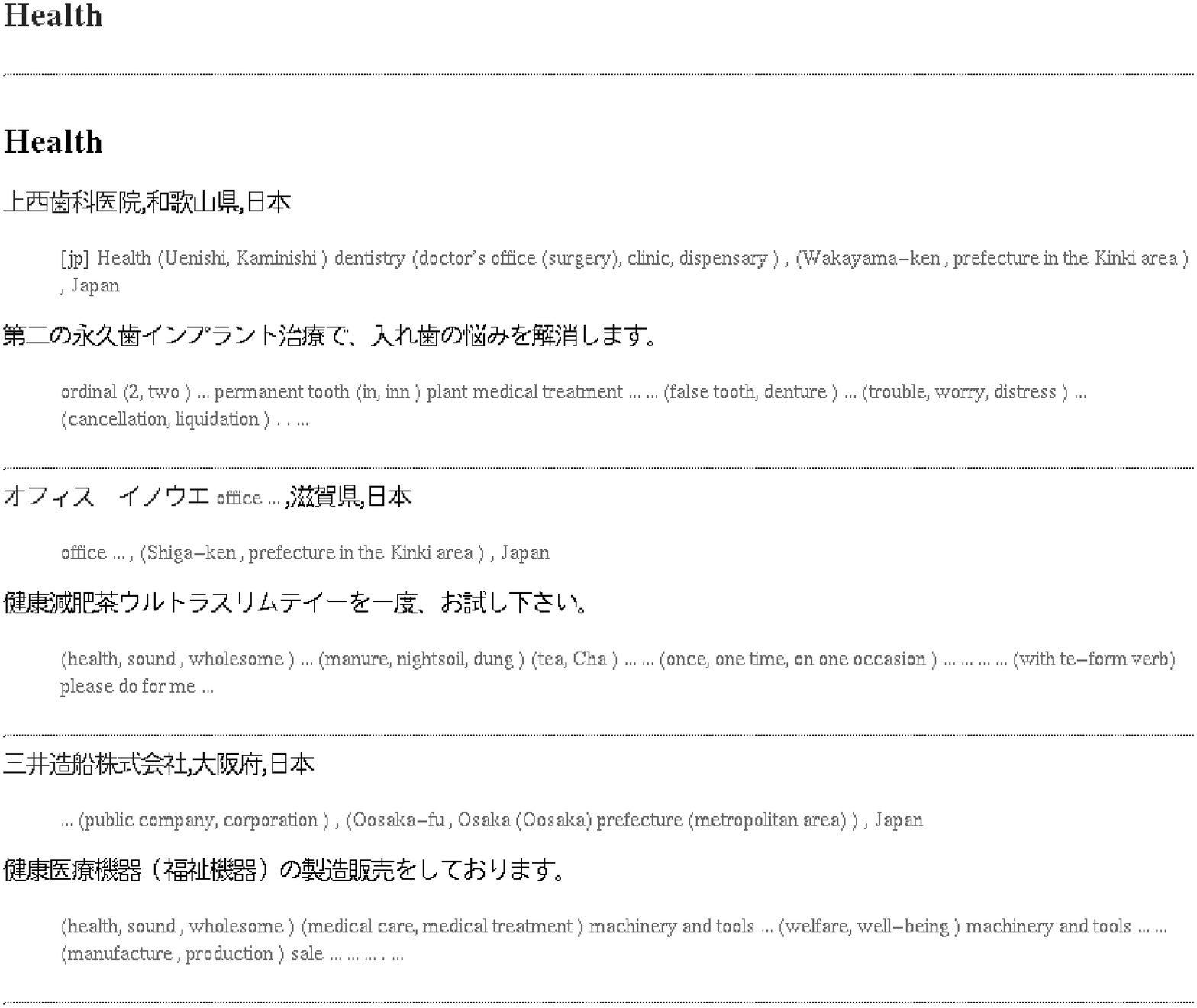,height=4in}}

}
\caption{Gisted items from Nihongo Yellow Pages
         \label{fig:nyp_gi}}
\end{figure*}

\begin{figure*}
\hbox{
\centerline{
\psfig{figure=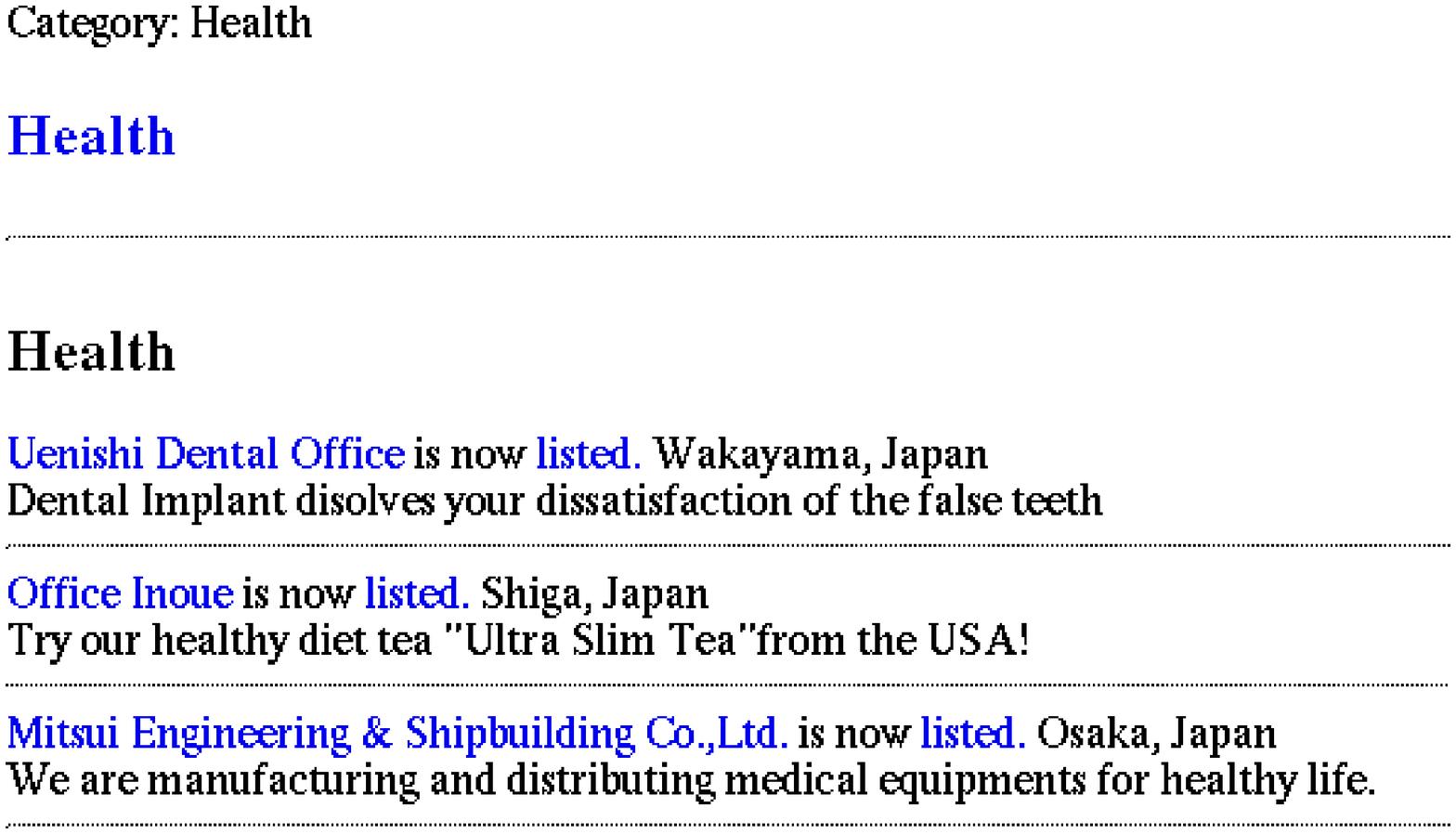,width=3.312in}}

}
\caption{Corresponding English items from Nihongo Yellow Pages
         \label{fig:nyp_en}}
\end{figure*}

Figure~\ref{fig:nyp_gi} shows the result of following this process for
the page in Figure~\ref{fig:nyp_jp}.  For comparison,
Figure~\ref{fig:nyp_en} shows the same entries as they appear in
an English version of the same business directory.

Our current implementation of the prototype handles gisting from
Japanese, French, and Spanish to English, though in this paper we
concern ourselves only with Japanese-English gisting.  Given the
simplicity of the approach, the main limiting factor in adding more
languages to the list is the availability of bilingual dictionaries,
though we expect that this problem may be ameliorated to some extent
by automatic algorithms for acquisition of bilingual lexicons
\cite{mel96b}.

\section{Evaluation Design Criteria}
\label{sec:methodology}
\label{sec:methodology:criteria}

The gisted text that appears in Figure~\ref{fig:nyp_gi} bears little
resemblance to an English translation of the Japanese content in
Figure~\ref{fig:nyp_jp}.  However, it does provide enough information
to support two critical {\em decisions} facing the user who has
arrived at that page:
\begin{itemize}
\item Deciding whether a link is worth following\ignore{Unfortunately,
links do not show up well in the figures.}
\item Deciding whether some text is worth having translated
\end{itemize}
A user interested in, say, podiatrists, can discern from the gisted
text in Figure~\ref{fig:nyp_gi} that the first entry in the Health
category is probably not worth navigating further.  Similarly, someone
interested in medical equipment manufacturers might well decide that
the third entry is worth translating, especially if they have a
particular interest in companies in Osaka.

The central issue of this paper is how to evaluate the extent to which
a gisting method helps the user to make decisions of this kind.  In
designing a methodology for answering that question, we were guided by
the following criteria:

\begin{quote}
{\bf Approximate real Web-based decision tasks.}  Since we have
characterized the role of gisting in terms of decision support, what
must be evaluated is the extent to which gisted material facilitates
decisions that resemble the choices available to the user when faced
with multilingual content on a Web page.  This consideration led us to
select a {\em categorization\/} paradigm, since both the real world
tasks involve a tradeoff between the time invested in assessing
relevance and the accuracy of the decision as well as the need to
select an appropriate action based on that assessment.

{\bf Minimize a priori biases.}  Users seeking information on the Web
are seldom given a pithy description of a topic by someone else.
Therefore it is important, in designing the experimental task, to
allow users to form their own internal characterization of a topic or
category, rather than pre-assigning category labels that incorporate
the experimenters' perceptions or biases.

{\bf Make the task easy to create.}  It is hoped that the methodology
proposed here can serve as an outline for other experimenters
investigating multilingual gisting, spoken language gisting,
translation, summarization, and related topics.  Therefore we aim for
an experimental design that requires little in the way of specialized
apparatus, preparation, and the like.
\end{quote}

The experimental design, adopting these criteria, is relatively
straightforward.  We define a task in which all subjects are faced
with the same categorization problem, but some of those subjects are
given materials in English to categorize while other subjects are
given the same {\em content\/} to categorize but in the form of gisted
text.  If the subjects given gisted materials make similar decisions
to the subjects given the English materials (allowing for normal
variability in people's judgments), we can conclude that the quality
of the gisting is reasonable.  The next section gives the details of
the experiment, including a way to assess the results quantitatively.

\section{Evaluation Study}
\label{sec:experiment}

\subsection{Materials}  

Experimental items were selected from the Nihongo Yellow Pages (NYP), a
business directory site on the World Wide Web \cite{nyp}.  The site
was chosen because it contains information across a variety of topic
areas, because each business directory listing consists of a concise
and informative description, and because most listings are available
in both Japanese and English.  In our experiments we used listings
from NYP's Education, Finance, What's New, Entertainment, and Health
categories, selecting a total of 73 business listings at random from
those areas.

For each of these listings we created a $3\times5$-inch index card
with a business advertisement in English and a corresponding card with
a ``gisted'' version of the same content as expressed in Japanese.  By
way of illustration, Figure~\ref{fig:nyp_en} shows three items in
English, with their corresponding gisted items appearing in
Figure~\ref{fig:nyp_gi}.

\subsection{Procedure}

\subsubsection{Creating Topical Categories}
\label{sec:procedure:piles}

In order to create topical categories in an objective way, we randomly
selected 32 of the 73 English cards and gave them to 3 different
subjects,\footnote{All subjects in this experiment were employees of
Sun Microsystems in Chelmsford, Massachusetts, solicited as
volunteers.  All were fluent in English and nobody who saw Japanese
materials was at all familiar with Japanese.}  with instructions to
sort the cards ``into 4-6 piles of roughly equal size, placing cards
in the same pile when you think they should 'go together', for example
because they are related to similar topics.''  One subject created 4
piles, another 6, and the third 7 piles.  We chose the 6 piles created
by the second subject as defining the topical categories for the
remainder of the study, noting that the topic distinctions made by the
three subjects were qualitatively similar overall.\footnote{As an
additional piece of information, we had each subject write a short
description of the topic for each pile, though those descriptions were
not used in the study.}

\subsubsection{Categorization Task:  The Control Condition}

\begin{figure*}
\hbox{
\centerline{
\psfig{figure=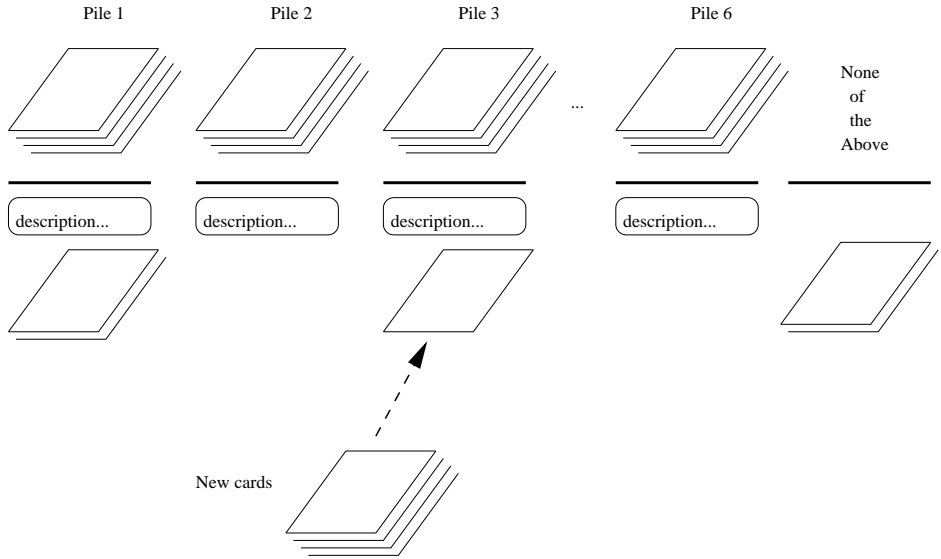,width=6in}}
}
\caption{Categorization of new items \label{fig:categorization}}
\end{figure*}

A set of 6 subjects participated in the control condition of the
experiment.  The procedure had two parts (see
Figure~\ref{fig:categorization}).

\begin{enumerate}

\item First, subjects were presented with the 6 piles of English cards
created as described above.  They were asked to read through each pile
and decide ``what you think each one is about.''  As a memory aid,
subjects were encouraged to write a description of their choosing on a
Post-It note for each pile, and place the note next to the
corresponding pile.

\item Having formed their own impression of the 6 topical categories,
subjects in the control condition were now given 32 new
randomly-selected cards in English.  They were instructed that for
each new card, they should decide in which of the 6 categories it
``belongs'' and place it next to the corresponding pile.  They were
also given the option of placing cards in a seventh ``none of the
above'' category.

\end{enumerate}
Subjects were told to take as long as they liked on both parts of the
categorization task, though Part 2 was timed for possible future use
of that information.

\subsubsection{Categorization Task: The Experimental Condition}

A set of 8 subjects participated in the experimental condition.  Part
1 of the experimental condition was completely identical to Part 1 of
the control condition: subjects looked at exactly the same 6 piles of
English cards and formed their own mental description of each topical
category, writing down a short description as a memory aid.

Part 2 was also identical, with one crucial exception: instead of
being given cards in English to place into categories, subjects were
given the corresponding {\em gisted Japanese\/} cards.

\subsubsection{Categorization Task: Random Baseline}

In order to obtain a lower bound for performance on this task, the
computer did 8 runs placing the gisted Japanese cards into the 7
categories at random.  We also computed lower bounds with the computer
making a forced choice, i.e. not allowing random selection to pick the
``none of the above'' category; the results differed negligibly.

\subsection{Analysis}

The categorization data gathered in the experiment were analyzed
following the method of Hripcsak et al. \cite{hripcsak1995}.  In their
study, they compared the performance of physicians, laypersons, and
several computer programs on the task of classifying chest radiograph
reports according to the presence or absence of 6 medical conditions.
Our adaptation of their analysis is almost completely direct, with
subjects in the control condition (English cards) corresponding to the
physicians, subjects in the experimental condition (gisted cards)
corresponding to laypersons, and each run of our random baseline
corresponding to a subject in their baseline conditions (simple
keyword-based classification).  

The basic idea in the analysis is to compute the ``distance'' between
subjects on the basis of their categorization behavior, and seeing
whether the average distance between an experimental subject and the
members of the control group is greater than the average distance of
control group members from each other.  We compute the distance
$d_{ijk}$ between two subjects $j$ and $k$ for experimental item $i$ as
the number of topical categories where the subjects disagreed for this
item, i.e. 0 if they placed item $i$ into the same category and 2 if
they did not.\footnote{This distance measure was used because Hripcsak
et al. included the more general case of allowing an item to be placed
into multiple categories, i.e. in their case distance could range from
0 to 6.}  The overall distance from subject $j$ to subject $k$ is then
just their average distance across all N items:
\begin{eqnarray}
d_{jk} & = & \sum_{i=1}^{N} d_{ijk}/N.
\end{eqnarray}

The main figure of interest in this study is how much the
categorization behavior of subjects in the experimental (gisted cards)
condition differs from behavior of subjects in the control (English
cards) condition.  The average distance from a gisted-card subject to
the English-card subjects is
\begin{eqnarray}
\bar{d}_{k} & = & \sum_{j=1}^{J} d_{jk}/J
\end{eqnarray}
where J is the number of English-card subjects.  The corresponding
average distance for English-card subjects is computed similarly,
though naturally the averaging excludes the distance of each subject
from himself or herself:
\begin{eqnarray}
\bar{d}_{l} & = & \sum_{1 \leq j \leq J, j \not= l} d_{jl}/(J-1).
\end{eqnarray}
Hripcsak et al. also give a method for computing confidence intervals
for these figures.  In addition they point out that the analysis holds
equally well for other inter-rater distance measures such as
Cohen's~$\kappa$, though they comment that in their study
Cohen's~$\kappa$ and the above distance measure produced essentially
the same results.

\subsubsection{Results}

\ignore{
Table~\ref{tbl:distance} shows, for each subject, the average distance
of his or her judgments from the judgments of the subjects in the
English-card (control) condition.\footnote{A graph with confidence
intervals will be given in the final paper.}
\begin{table}
\begin{center}
\begin{tabular}{ccc}
\hline
English & Gisted & Random \\ 
\hline
0.60 &   0.74 &   1.64 \\
0.63 &   0.77 &   1.68 \\
0.66 &   0.86 &   1.70 \\
0.74 &   0.90 &   1.70 \\
0.91 &   1.02 &   1.75 \\
1.06 &   1.13 &   1.78 \\
     &   1.21 &   1.80 \\
     &   1.22 &   1.81 \\
\hline
\end{tabular}
\end{center}
\caption{Average distance from subjects in the English-card condition
         \label{tbl:distance}}
\end{table}
}

\begin{figure}
\hbox{
\centerline{
\psfig{figure=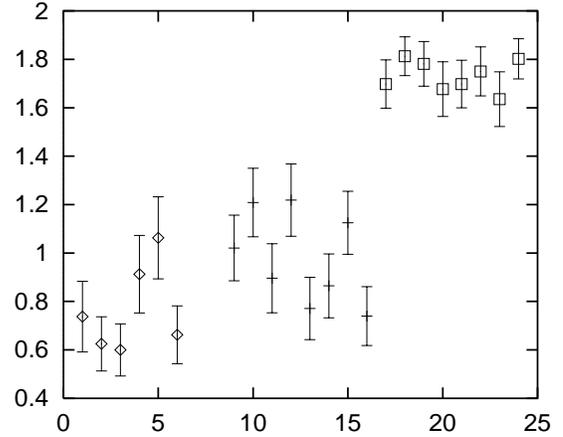,width=3.312in}}
}
\caption{Left to right: English condition, Gisted condition, Random condition
  \label{fig:graph}}
\end{figure}

Fig~\ref{fig:graph} shows, for each subject, a point (and 95\%
confidence interval), representing its distance on average from the
judgments of the subjects in the English-card (control) condition.
(Recall that distances range from 0 to 2.)  As one should expect, the
categorization behavior of subjects given degraded information (gisted
cards) is far closer to the control group than random choice, but
generally appears appears to differ from that of subjects in the
control group, who were given full information in the form of English
cards.

We plan to replicate the study with a greater number of subjects, in
order to better assess the significance of the variability that
appears within the control group --- in particular, whether the degree
of variance in the control group, suggested by comparatively greater
distances for the 4th and 5th subjects, will turn out to be present or
not given a larger sample.  In addition, it has been suggested that an
additional, informative control in this experiment would be a group
that performed the experiment using cards entirely in Japanese (for
both the topical ``piles'' and the cards to be categorized); the
materials for this condition are easily created, but our ability to
perform the experiment will depend upon the availability of subjects
who are fluent in Japanese.

\section{Discussion}

Our central concern in this paper is not the method used for gisting
--- though of course that is also of interest --- but rather the
evaluation methodology we have designed.  Were we to extend the
gisting prototype, for example by improving dictionary coverage,
adding automatic disambiguation, or manipulating word order, the value
added by those changes could be measured simply and effectively by
adding a condition to the above experiment in which subjects received
cards with the putatively improved information.  Similarly, anyone
else's method for conveying the content of Japanese Web pages
(e.g. Temple, \cite{vanni:gisting}) can be evaluated in terms of its
value for gisting (i.e. decision support) simply by creating the
corresponding materials from the same Japanese items we used to
produce gisted cards in our experiment.  If one method for producing
gists is better than another, then subjects given that information
should behave closer to the ``ideal'' case (defined here by the
behavior of subjects who receive information in English), as assessed
quantitatively by the distance measure.  Additional measures might
also be brought into play, such as a comparison of the time it takes
to make decisions given variant forms of information, or differences
in the time-accuracy tradeoff that results when time limitations are
imposed.

The evaluation methodology we have proposed generalizes easily to any
number of other tasks that have similar characteristics, namely
domains in which restricted or alternate-form information is used in support
of a decision-making because of limits on time, space, or user
knowledge.  Some examples:
\begin{itemize}
\item In environments where text summarization is used to decide the
      disposition of full documents, e.g. routing of memoranda or
      scientific articles, this methodology could be used to evaluate the
      quality of summaries.
\item In environments where key elements are extracted from a stream
      of speech input, e.g. automatic monitoring of radio traffic,
      this methodology could be used to evaluate the extraction
      technology.
\item In environments where decisions are made on the basis of
      text-to-speech output, e.g. spoken language interfaces,
      this methodology could be used to evaluate the clarity
      of the speech synthesizer.
\item In environments where alternative versions of text or images
      can be presented, e.g. the selection of Web-based advertising
      based on client bandwidth, this methodology could be used to
      assess the impact of the advertisement format on users'
      interest level.
\end{itemize}

We will be happy to make our experimental materials available to
other researchers on request.

\section{Acknowledgements}

This research was conducted at Sun Microsystems Laboratories in
collaboration with Gary Adams.  The author also gratefully
acknowledges the contributions of Mark Torrance and Bob Kuhns in
development of the prototype, helpful discussions on experimental
method with Gary Marchionini, Gail Mauner, and Nicole Yankelovich,
assistance by Cookie Callahan in preparing experimental materials, and
the time of Sun Microsystems volunteers who participated in the
experiment.

\bibliographystyle{aaai}
\bibliography{gisting,general,learning,distrib,nlstat,ibm_master}

\end{document}